\documentclass[]{fairmeta}
\usepackage{amsthm,amsmath,amssymb,amsfonts}
\usepackage{graphicx}
\usepackage{natbib}
\usepackage{booktabs}
\usepackage{tabularx}
\usepackage{makecell}
\usepackage{algorithm}
\usepackage{algorithmic}
\usepackage{array}
\usepackage{textcomp}
\usepackage{stfloats}
\usepackage{url}
\usepackage{verbatim}
\usepackage{multirow}
\usepackage{bm}
\usepackage[table]{xcolor}
\usepackage{float}
\usepackage[misc]{ifsym}
\usepackage{subcaption}

\newtheorem{theorem}{Theorem}[]

\newtheorem{remark1}[theorem]{Remark}

\newcommand{\modelname}{Directing the World}

\title{\modelname{}: Fast Autoregressive Video Generation with Compositional Human-Camera Control}

\author[1]{Haoyuan Wang\textsuperscript{*}}
\author[1]{Yabo Chen\textsuperscript{*}}
\author[1]{Haibin Huang}
\author[1]{Chi Zhang}
\author[1]{Xuelong Li\textsuperscript{\Letter}}

\affiliation[1]{Institute of Artificial Intelligence, China Telecom (TeleAI)}

\contribution{\textsuperscript{*}Equal Contribution}

\metadata[Corresponding Author]{Xuelong Li (\email{xuelong$\_$li@ieee.org})}

\begin{document}

\abstract{Building interactive world models requires not only generating visually realistic videos, but also simulating controllable real-world dynamics over long temporal horizons. In practice, autoregressive video generation provides a scalable foundation for such dynamic world simulation, where future observations are progressively synthesized from previously generated context. However, maintaining temporal stability over extended rollouts remains fundamentally challenging, as small prediction errors can accumulate and quickly degrade long-horizon generation. This challenge becomes even more pronounced when the model must support heterogeneous controls, such as human motion and camera trajectories, since different control signals may interfere with each other and destabilize the pretrained video prior. Moreover, existing controllable video generation methods often face a difficult trade-off between generation quality and fine-grained controllability.To address these challenges, we present \textit{``Directing the World''}, a fast autoregressive framework for controllable world-model video generation with compositional human-motion and camera-trajectory control. Our key insight is to decouple the learning of dynamic control factors while preserving a unified autoregressive video prior. Specifically, we introduce a ``Fast--Slow Memory training strategy'' that stabilizes long-horizon rollout learning and improves convergence during controllable post-training. To enable accurate and temporally smooth human-motion control, we further design a $t$-guided Dynamic Projection mechanism together with a refined Motion-CFG strategy, which injects motion conditions adaptively across denoising stages and improves motion alignment without sacrificing visual fidelity. This design also naturally supports multi-person motion control within the same video, including simultaneous control of multiple people and precise control of an individual subject among multiple people.
After acquiring a robust human-motion prior, we introduce camera-trajectory control in a second stage, allowing the model to compose human dynamics with camera motion for controllable world exploration. To support this training process, we construct a large-scale controllable world-model dataset with synchronized video, text, human-motion, and camera-trajectory annotations, and further organize it into motion-centric and camera-centric subsets for decoupled two-stage learning.
Extensive experiments demonstrate that our method achieves stable long-horizon generation, precise human-motion control, and coherent camera-controlled world dynamics while maintaining high visual quality. See more at https://whydahuzi.github.io/Directing-the-World.github.io/.

\vspace{0.5em}
}

\maketitle

\begin{figure}[t]
  \centering
  \includegraphics[width=1\linewidth]{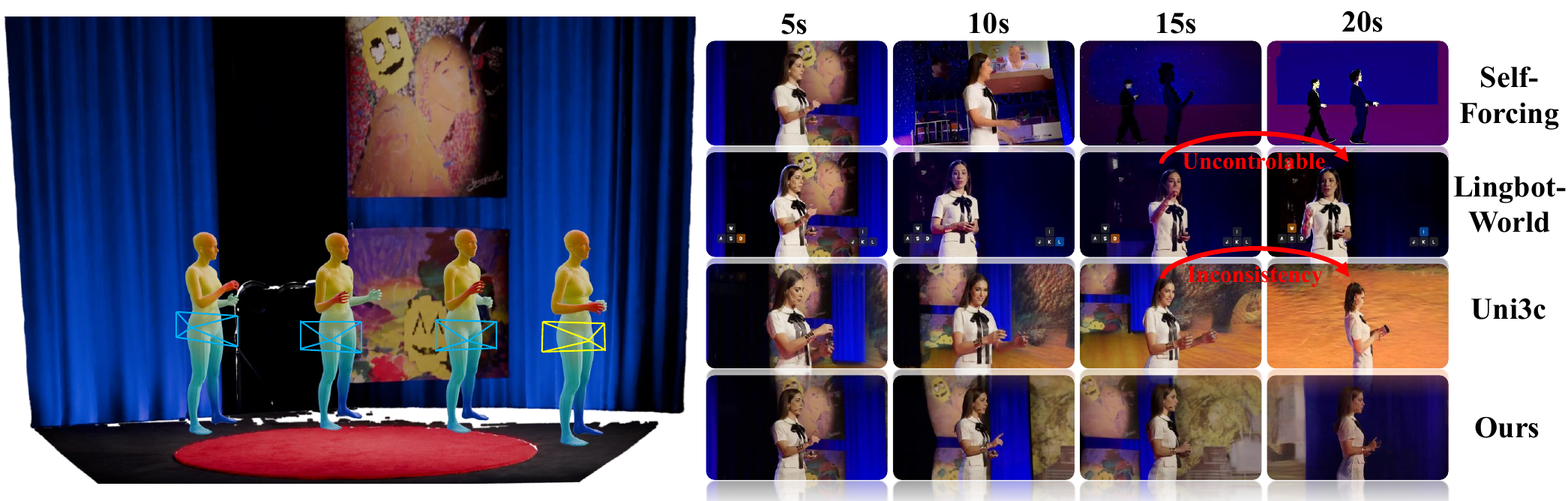}
  \caption{Illustration of a long-horizon world exploration task with controlled human motion. Given an input image, the motion of a female speaker, and a prescribed camera trajectory, existing methods still struggle to generate long-sequence world-model videos with fine-grained joint control. They either fail to maintain long-range temporal coherence or provide only coarse human-motion control, often limited to simple translational movements.
  Self-Forcing~\cite{selfforcing} suffers from long-term memory degradation, Uni3C~\cite{uni3c} produces inconsistent scene evolution during sustained world exploration, and Lingbot-World~\cite{gao2026lingbotworld} remains largely uncontrollable for precise human motion, often generating a foreground person detached from the scene. In contrast, our \textbf{Directing the World} enables stable long-video generation with consistent human-motion control and coherent camera-based world exploration.}
  \label{fig:teaser}
\end{figure}

\section{Introduction}
\label{sec:intro}

Simulating complex real-world dynamics is a crucial step toward interactive video world modeling and controllable long-horizon generation~\cite{wan,causvid,selfforcing,mmpl,genie2,relic,live,astra,teleworld,interactivewm,ldt_tmm2026,an2026aiflow,shao2026aiflownetworkedge,
zhang2026symphomotion,zhang2026telephysics,huang2025zero}. Recent video-based world models have advanced from passive visual synthesis to controllable simulation, enabling applications in embodied agents, virtual reality, game engines, and digital content creation~\cite{eccv_domainfusion,NEURIPS2025_DDPTM,chen2024liftimage3d,chen2024cascadezero123,chen2026full4dgeneratingfullscope4d,xiang2026videoweaveunlockinggeometricconsistency}.
However, most existing methods primarily model the evolution of the surrounding world, such as scene consistency, viewpoint changes, and long-term appearance preservation, while treating humans as passive visual elements rather than controllable active agents. Since human activities are central to real-world environments, an interactive world model should support not only persistent scene evolution, but also precise and compositional control over human agents within the same dynamic world. This requirement becomes even more important in multi-agent scenarios, where multiple people should move, interact, and remain spatially consistent in a shared environment~\cite{customvideo_tmm2026}.

Existing human-centric video generation methods, including pose-guided animation, motion transfer, character animation, and motion generation~\cite{realisdance_dit,wan_move,mimicmotion,motionflow_tmm2026}, provide useful tools for controlling individual human motion, but they are usually designed for short clips or localized editing. As a result, they often lack the long-horizon memory required by world models, leading to identity drift, body inconsistency, jitter, flickering, and degraded scene coherence during extended roll-outs. Meanwhile, recent controllable world-generation methods~\cite{cameractrl,motionctrl,cami2v,uni3c,cameractrl2} improve scene-level exploration and viewpoint consistency, but their control interfaces remain largely coarse-grained and are insufficient for specifying fine-grained articulated human dynamics, especially when multiple agents coexist in the same world.

Therefore, video-based world modeling requires more than visually plausible long-video generation. It calls for a unified autoregressive framework that preserves world memory over long horizons while supporting controllable human dynamics as part of the evolving environment. Such a framework should enable stable simulation under heterogeneous controls, maintain spatial coherence under large viewpoint changes, and support both single-agent precision and multi-agent coordination within a persistent world.

To surmount the aforementioned challenges, we propose \textbf{Directing the World}, an autoregressive video world model for stable long-horizon generation under joint human-motion and camera-trajectory control. Our framework consists of three coupled components: (1) an autoregressive generation backbone built upon MMPL~\cite{mmpl} to preserve long-term world memory and temporal consistency; (2) an SMPL condition injection framework for controllable human dynamics; and (3) a causal camera-control framework for coherent world exploration under large camera motion.

To address instability, condition entanglement, and controllability-quality trade-offs, we introduce a multi-stage training framework with decoupled data filtering. Specifically, we first train on motion-centric samples with static backgrounds to learn controllable human dynamics, and then introduce dynamic camera trajectories and world-exploration signals. This staged design reduces interference between heterogeneous controls while preserving the strong long-video prior of the autoregressive base model.

To stabilize controllable post-training, we introduce a ``Fast-Slow Memory'' training paradigm, where different modules adapt at different rates. This strategy preserves the intrinsic generative capability and long-horizon memory of the pretrained world model while allowing newly introduced control modules to effectively learn human-motion and camera-trajectory conditions.

For precise human-motion control, we propose a \textbf{timestep-guided ($t$-guided) projection mechanism}. Instead of injecting SMPL conditions in a timestep-agnostic manner, our method modulates motion features according to the current denoising stage. This enables strong motion guidance when it is most effective while avoiding over-dominant condition injection that may cause jitter, body twitching, or unstable artifacts. As a result, the model follows the input motion more faithfully while preserving visual realism and temporal smoothness. Moreover, this mechanism naturally supports flexible multi-person control, including simultaneous control of multiple people and fine-grained control of a target individual among multiple subjects.

Overall, \textbf{Directing the World} enables decoupled yet harmonious control over human dynamics and world-level camera exploration, achieving stable, temporally coherent, and high-quality autoregressive world simulation.

In summary, our main contributions are as follows:
\begin{itemize}
\item We propose a unified autoregressive world-modeling framework that decouples human-motion and camera-trajectory control while integrating them coherently within a single generative model.

\item We introduce a progressive multi-stage training strategy and a ``Fast-Slow Memory'' training paradigm to stabilize controllable post-training, reduce interference between heterogeneous controls.

\item We design a timestep-guided ($t$-guided) projection mechanism for SMPL condition injection, enabling faithful and smooth human-motion control across denoising stages, which also supports simultaneous multi-person control.

\item We establish a complete data cleaning and filtering pipeline for motion-camera controllable world-model training, and will release curated datasets.
\end{itemize}

\section{Related Work}
\label{sec:related}

\subsection{Autoregressive and Interactive World Models}
Recent progress in video generation has increasingly pushed world modeling beyond short-horizon visual synthesis toward interactive, persistent, and controllable simulation. Rather than producing isolated video clips, emerging world models aim to generate future observations conditioned on historical context and external user inputs, enabling agents or users to navigate, act, and interact within generated environments. This direction has been explored in navigation-conditioned visual prediction, real-time interactive world generation, open-source video-based simulators, and multimodal 4D world modeling~\cite{bar2025navigation,genie3,gao2026lingbotworld,teleworld}. More recently, video world models have also been extended toward physical-AI simulation, autonomous-driving scenario generation, robot control, and unified 4D world-action modeling~\cite{drivinggen2026,lingbotva2026,xwam2026,an2026aiflow,shao2026aiflownetworkedge}. Together, these studies suggest a broader shift from passive video synthesis to interactive world simulation, where temporal memory, causal generation, real-time response, and controllable world evolution become increasingly important.

Despite these advances, the controllability of existing world models remains largely coarse-grained. Most systems are driven by text prompts, navigation actions, keyboard or mouse inputs, camera poses, or promptable scene events. These controls are effective for specifying agent movement, viewpoint changes, or high-level scene events, but they provide limited support for fine-grained articulated human dynamics. In particular, body poses, motion styles, limb coordination, hand-object interactions, temporally ordered human actions, and coordinated multi-person motion remain difficult to control reliably, especially over long autoregressive rollouts. This limitation becomes more critical in multi-agent world modeling, where multiple humans should be independently or jointly controlled while moving and interacting within the same persistent environment. Therefore, while recent world models have made substantial progress in interactivity, real-time response, long-term memory, and scene-level controllability, their control interfaces are still not well suited for specifying fine-grained human activities and multi-person dynamics in dynamic scenes.

\subsection{Human Motion-Controlled Video Synthesis}
Controllable human video generation has advanced from coarse pose-guided animation to more structured and geometry-aware motion control. Early methods typically use 2D keypoints, dense pose maps, or skeletal heatmaps as conditions, which provide effective coarse guidance but are limited for complex 3D articulated motions with self-occlusion, large pose changes, or viewpoint variation~\cite{animateanyone,mimicmotion}. Recent methods introduce stronger human priors, such as SMPL, depth, and 3D-aware body representations~\cite{smpl,champ,realisdance_dit,uni3c,mimicmotion,animateanyone}, to improve geometric faithfulness and motion plausibility. For example, RealisDance-DiT adapts large video foundation models for in-the-wild character animation, Uni3C enhances motion control with explicit 3D representations~\cite{realisdance_dit,uni3c}, and Wan-Move explores trajectory-level motion control on top of the Wan family~\cite{wan_move}.

Despite these advances, most existing methods are designed for diffusion-based generation, short clips, image animation, or subject-specific settings. Motion conditions are often injected through direct concatenation, feature addition, or ControlNet-style branches, which may not align with the timestep-dependent latent distribution of video diffusion models. When extended to long-horizon generation, strong motion conditions can therefore cause flickering, identity drift, or noisy artifacts.

\subsection{Camera Control in Video Generation}
Camera control is another key component of controllable video synthesis. Recent methods use explicit pose parameterizations, such as camera extrinsics, Pl\"ucker rays, or viewpoint tokens, to guide camera-conditioned video diffusion and 3D-aware generation~\cite{cameractrl,motionctrl,cami2v,uni3c,cameractrl2}. CameraCtrl introduces plug-and-play camera control for text-to-video diffusion, MotionCtrl studies unified camera and object-motion control, and CamI2V improves camera-controlled image-to-video synthesis with geometry-aware modeling~\cite{cameractrl,motionctrl,cami2v}. Uni3C is particularly relevant as it unifies camera and human-motion control through explicit 3D enhancement and dedicated camera feature extraction~\cite{uni3c}.

However, camera control requires preserving scene identity while updating viewpoint-dependent content, making it highly sensitive to temporal inconsistency and structural collapse. Moreover, existing camera-control methods are mainly developed for non-autoregressive or globally conditioned generation, where the full temporal context is available during synthesis. Directly applying such global camera-conditioning modules to block-wise autoregressive generation creates a mismatch between globally encoded camera features and causally generated temporal windows.

\section{Method}
\label{sec:method}

\begin{figure}[t]
  \centering
  \includegraphics[width=1\linewidth]{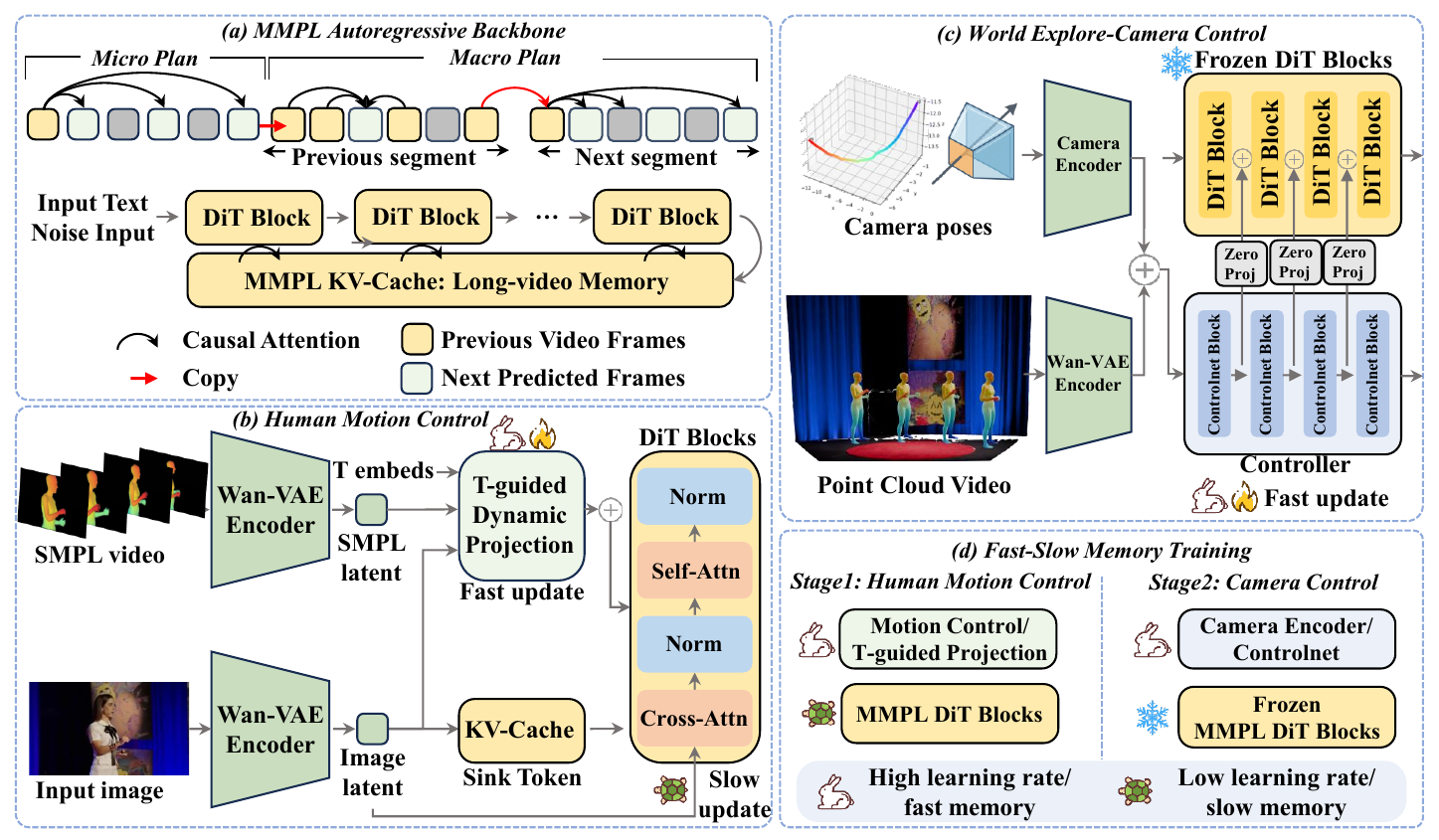}
\caption{
Overview of our controllable autoregressive video generation framework.
(a) \textbf{MMPL Autoregressive Backbone.} We adopt an MMPL-based autoregressive backbone for long-horizon video synthesis.
(b) \textbf{Human Motion Control.} SMPL videos are injected as structural guidance for fine-grained human-motion control.
(c) \textbf{World Explore-Camera Control.} Given the input image and camera trajectory, we render a first-frame point-cloud video as camera-aware guidance for controllable scene exploration.
(d) \textbf{Fast-Slow Memory Training.} Both control branches are integrated into the backbone via Fast-Slow Memory Training, preserving generation capability while enabling efficient controllable adaptation.
}
  \label{fig:pipeline}
\end{figure}

\subsection{Overview}
\label{sec:method_overview}

As discussed in Section~\ref{sec:intro}, controllable video world modeling requires stable long-horizon autoregressive memory, fine-grained human-motion control, and coherent camera-based world exploration. Directly combining these heterogeneous controls, however, can easily introduce condition interference, temporal instability, and degraded generation quality.

To address this, we propose \textbf{Directing the World}, a two-stage autoregressive ControlNet-based framework for controllable video world modeling. Our design follows a ``decouple then integrate'' principle: we first learn human-motion control under relatively static camera settings, and then progressively introduce camera-trajectory control on top of the acquired motion prior. This staged formulation allows different control signals to be learned separately while being integrated into a unified autoregressive world model.

As shown in Figure~\ref{fig:pipeline}, our framework contains four key components. First, we adopt MMPL~\cite{mmpl} as the long-horizon autoregressive backbone, as introduced in Section~\ref{sec:preliminaries}. Second, we build a fine-grained motion-control branch based on SMPL sequences, together with a timestep-guided projection mechanism that adapts geometric motion conditions to timestep-dependent denoising latents, as described in Section~\ref{sec:motion_control}. Third, we redesign the camera-control pathway to align with block-wise autoregressive generation, enabling global trajectory understanding while injecting temporally matched control features into each generation block, as discussed in Section~\ref{sec:camera_control}. Finally, we introduce a ``Fast-Slow Memory'' training strategy to preserve the long-horizon generative prior of the pretrained backbone while allowing newly added control modules to adapt efficiently, as detailed in Section~\ref{sec:memory_arch}.

Together, these designs enable stable long-horizon world simulation with precise human-motion control and coherent camera-based exploration.

\subsection{Preliminaries}
\label{sec:preliminaries}

We first introduce the autoregressive video generation framework adopted in our method. Our framework builds upon MMPL~\cite{mmpl}, a long-horizon autoregressive video generation model designed for efficient and temporally consistent world simulation. Unlike full-sequence video diffusion models, MMPL generates videos in a block-wise autoregressive manner, which improves generation efficiency while maintaining long-range temporal consistency.

\paragraph{MMPL-based Long Video Autoregressive Generation}
\label{sec:mmpl_prelim}
Standard autoregressive video generation~\cite{videopoet,phenaki,nova,videomar,ca2vdm,far} typically follows a causal formulation $p(x_t \mid x_{<t})$, where each frame is generated only from past observations. While effective for sequential synthesis, this unidirectional dependency can accumulate errors over long horizons, leading to temporal drift and structural inconsistency.

MMPL addresses this issue with a \textit{Plan-then-Populate} strategy. For each video segment, instead of generating frames strictly one by one, MMPL first predicts a small set of planning frames from the initial frame $x_{start}$, including the terminal frame $x_{end}$:
\begin{equation}
    p(\mathcal{M} \mid x_{start}) = p(x_{early}, x_{mid}, x_{end} \mid x_{start}) .
\end{equation}
These planning frames provide high-level temporal anchors for the segment. In particular, the terminal frame $x_{end}$ serves as a future anchor that constrains the generation of intermediate frames together with the initial frame $x_{start}$. As a result, intermediate-frame synthesis is guided by both past and future context, rather than relying only on historical information.

Moreover, $x_{end}$ is reused as the initial condition of the next segment, thereby connecting consecutive segments into a coherent long video. This design reduces cross-segment drift and enables MMPL to maintain stable world appearance and temporal structure over extended autoregressive rollouts.

\subsection{Fine-grained Human Motion Control}
\label{sec:motion_control}
Existing long-video generation frameworks generally lack the ability to achieve precise human motion control. Some methods~\cite{video_xfun,dwpose} represent human motion using DWPose-style 2D keypoints, which provide only sparse image-plane constraints and therefore suffer from inherent ambiguity in 3D body orientation, especially when distinguishing front-facing and back-facing poses. Moreover, since 2D keypoints do not explicitly encode body shape, limb length, or human-scale priors, directly injecting such conditions into long-video generation can easily produce distorted characters, inconsistent body proportions, and unnatural limb deformation across time.

\begin{figure}[t]
  \centering
  \includegraphics[width=1\linewidth]{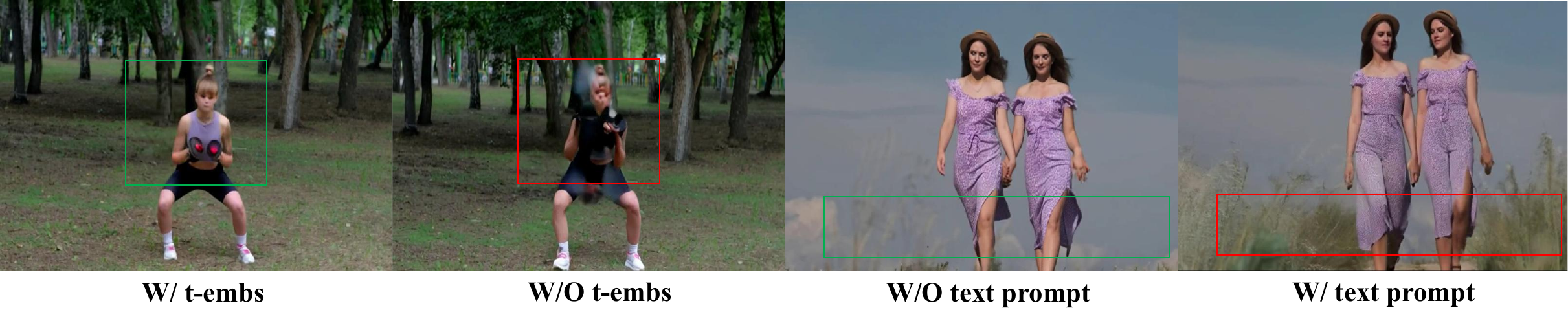}
   \caption{Effect of $t$-embs and prompt specificity on human-motion control. Timestep embeddings stabilize condition injection, reducing motion jitter and background flickering. Detailed prompts may introduce semantic motion biases that compete with the input condition, while a fixed prompt, together with Motion-CFG, helps the model follow the provided motion more faithfully.}
  \label{fig:compar1}
\end{figure}

To address these issues, we introduce SMPL as an explicit 3D human representation for motion guidance. However, SMPL is defined in 3D geometric space, while video generation operates in the diffusion latent/pixel space, leading to a cross-modal alignment problem. We therefore perform latent-space alignment to bridge SMPL geometry and video generation features. Meanwhile, direct motion-condition injection often causes background flickering and human jitter, especially under strong control signals. To ensure fine-grained yet stable control, we introduce a timestep-guided projection mechanism that adapts the strength and distribution of the motion condition to different denoising stages. In addition, as discussed in the preliminary, existing AR image-to-video models are typically built upon T2V backbones with sink tokens, whose cross-attention layers tend to over-emphasize textual prompts and weaken motion guidance, resulting in inaccurate or distorted human motion. To mitigate this guidance-modality mismatch, we further adopt a Motion-CFG strategy that strengthens SMPL-driven motion adherence while preserving the semantic prior of the base model.

\begin{figure}[t]
  \centering
  \includegraphics[width=0.85\linewidth]{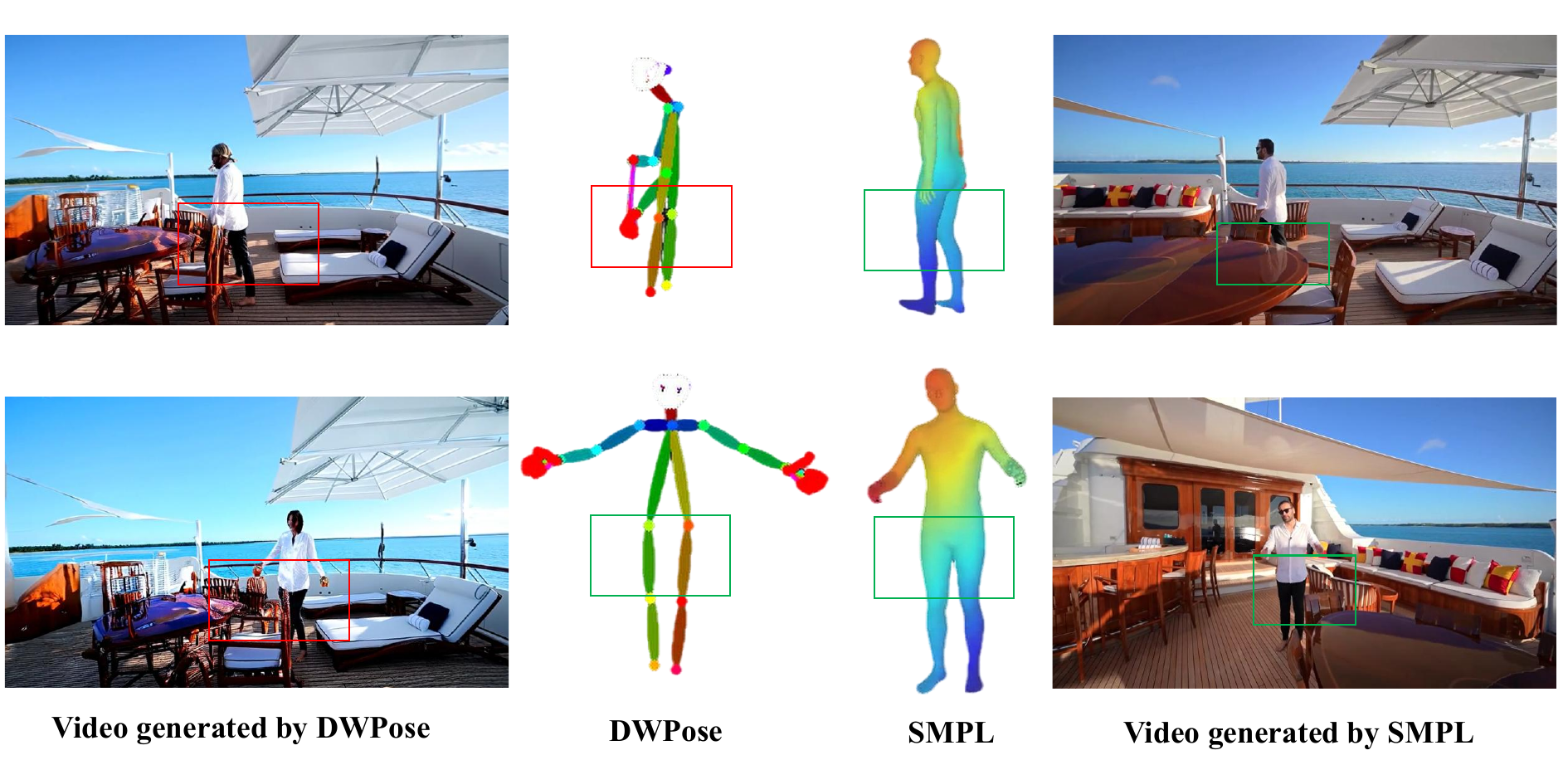}
   \caption{Comparison between DWPose and SMPL-based motion conditions. Compared with DWPose keypoints, SMPL provides stronger 3D human-body priors by explicitly modeling body pose, shape, and articulated skeletal structure. When used as the motion condition, such structured 3D priors help preserve plausible body geometry during condition injection, reducing motion distortion, limb twisting, and pose inconsistency. As a result, SMPL-based conditions enable more stable and faithful human-motion control.
}
  \label{fig:compar2}
\end{figure}

Then, we detail our solution comprising latent space alignment, timestep-guided projection, and a refined Motion-CFG strategy.

\vspace{0.5em}
\noindent\textbf{Latent Space Alignment.}
To bridge the gap between 3D human geometry and video latent generation, we first spatially align the SMPL body with the target human in the video. Specifically, we optimize the scale and spatial bounding-box overlap between the rendered SMPL sequence and the target person region, ensuring that the projected body structure remains geometrically consistent with the visual subject across frames. This alignment step significantly reduces discrepancies in body scale and spatial position before condition injection.

After alignment, the rendered SMPL sequence is encoded using the same VAE encoder as the video generation backbone, mapping the 3D motion condition into the same low-dimensional latent manifold as the denoising video latent. This unified latent-space representation alleviates the feature-distribution mismatch between geometric control signals and diffusion video features. In practice, directly injecting raw 3D conditions into autoregressive diffusion generation often leads to unstable optimization, slow convergence, or even training collapse due to severe modality inconsistency. By compressing both modalities into a shared latent representation space, our latent-space alignment strategy enables more stable training and more coherent motion-conditioned video synthesis.

\vspace{0.5em}
\noindent\textbf{$T$-guided Dynamic Projection.}
Although latent-space alignment reduces the modality gap between SMPL geometry and diffusion video features, directly injecting motion conditions into the noisy latent still introduces severe instability during autoregressive denoising. As visualized in Fig.~\ref{fig:compar1}, naive condition fusion can disrupt the original noise distribution, leading to background flickering, motion jitter, and structural inconsistency across long temporal horizons.

Let $z_0$ denote the clean video latent, and let $z_t=\alpha_t z_0+\sigma_t\epsilon$, $\epsilon\sim\mathcal{N}(0,I)$, denote its noisy version at diffusion timestep $t$, where $\alpha_t$ and $\sigma_t$ control the signal-noise ratio. Since early timesteps are dominated by noise while later timesteps gradually approach the clean video manifold, the latent distribution is inherently timestep-dependent.

Let $c_{\mathrm{smpl}}$ denote the latent feature encoded from the rendered SMPL sequence. A naive strategy directly injects motion guidance as $z_t^{\mathrm{naive}}=z_t+\alpha c_{\mathrm{smpl}}$, where $\alpha$ is a fixed scaling factor. However, this timestep-invariant fusion ignores the evolving statistics of $z_t$, while $c_{\mathrm{smpl}}$ remains in a relatively clean and stable latent regime. Consequently, the relative perturbation $r_t=\|\alpha c_{\mathrm{smpl}}\|/\|z_t\|$ varies significantly across denoising stages. In high-noise timesteps, the motion condition can be overwhelmed by stochastic noise and become ineffective; in low-noise timesteps, the same condition may become excessively strong and distort the reverse diffusion trajectory. This timestep mismatch is one of the primary causes of unstable controllable generation in long autoregressive videos.

To address this issue, we introduce a timestep-aware projection module $\tilde{c}_t=\mathcal{P}(c_{\mathrm{smpl}},t)$ and inject the projected condition as
$$
z_t^{\mathrm{input}} = z_t + \alpha \tilde{c}_t .
$$
Instead of directly superimposing static motion conditions onto noisy latents, the proposed projection module dynamically adapts the scale and feature distribution of the control signal according to the current denoising timestep. This allows the motion guidance to remain both effective and non-destructive throughout the diffusion process.

Intuitively, our $t$-guided projection performs coarse-to-fine motion control across denoising stages. At high-noise timesteps, the model mainly focuses on robust global motion structure and large body dynamics; at low-noise timesteps, the projection gradually shifts toward finer geometric details and appearance-consistent motion refinement. By explicitly matching motion conditions to timestep-dependent latent distributions, the proposed design significantly improves temporal stability and controllable generation quality in long-horizon autoregressive world modeling.

\vspace{0.5em}
\noindent\textbf{Motion CFG Strategy.}
Since many autoregressive image-to-video frameworks adopt sink-token-based text-to-video architectures for TI2V generation, the final synthesis tends to be highly influenced by textual descriptions. However, in our setting, human dynamics should be primarily governed by the injected SMPL guidance rather than dense text prompts. In practice, we observe that overly strong textual conditioning can weaken the effect of SMPL signals, resulting in inaccurate motion following or even motion-control failure.

To counteract this ``text-dominance'', we adopt a Motion-CFG strategy: fixing the text prompt and using the SMPL condition for classifier-free guidance. The guidance formula is $\epsilon = \epsilon_{uncond} + s \cdot (\epsilon_{cond} - \epsilon_{uncond})$, where the condition is the SMPL input.
Crucially, for the unconditional branch $\epsilon_{\mathrm{uncond}}$, standard CFG implementations often use a zero-filled tensor as the condition input. In our setting, however, a purely zero-valued latent may be less compatible with the color-space statistics of the video latent and can be inadvertently confused with the SMPL-conditioned representation. We therefore encode a black video using the same VAE encoder and use the resulting latent as the unconditional input. This choice provides a visually neutral yet latent-compatible reference condition for Motion-CFG, making the unconditional branch easier to distinguish from the SMPL control branch in the color/appearance space. Empirically, this design tends to produce a more stable guidance direction and improves SMPL motion adherence under strong textual priors.

\subsection{Causal-Aligned Camera Control}
\label{sec:camera_control}

Inspired by Uni3C~\cite{uni3c}, we introduce a causal-aligned camera control module that represents camera poses with Pl\"ucker coordinates and adapts camera-control injection to the block-wise autoregressive generation paradigm. Directly applying Uni3C to our framework is suboptimal, since Uni3C encodes the entire camera sequence with global self-attention, whereas MMPL generates videos progressively in temporal blocks. Injecting globally aggregated camera features into each block may therefore conflict with the causal structure expected by the autoregressive base model.

To address this mismatch, we decouple \emph{camera feature encoding} from \emph{camera feature injection}. During training, the full ground-truth camera trajectory is available, so we retain Uni3C's global self-attention encoder to capture long-range trajectory dependencies and temporal smoothness. Given a camera trajectory $C=\{c_1,c_2,\dots,c_T\}$, the camera encoder produces contextualized features
\begin{equation}
    H=\{h_1,h_2,\dots,h_T\}=\mathcal{E}_{\mathrm{cam}}(C).
\end{equation}

During autoregressive generation, MMPL synthesizes the video in temporal windows. For the $i$-th generation block with frame indices $\mathcal{W}_i$, we inject only the temporally aligned camera features
\begin{equation}
    H_i=\{h_t \mid t\in\mathcal{W}_i\}
\end{equation}
into the denoising network, rather than using the entire feature sequence $H$. In this way, the camera branch still benefits from globally contextualized trajectory representations, while the base generator receives only block-local control signals. This preserves temporal alignment and makes camera control compatible with block-wise autoregressive video generation.

\subsection{Fast-Slow Training for Stable Adaptation}
\label{sec:memory_arch}

After introducing motion and camera controls into the pretrained autoregressive world model, the key challenge is to adapt the new control modules while preserving the base model's long-video generation capability and spatiotemporal consistency. Directly fine-tuning all parameters under heterogeneous conditions can cause unstable optimization and modality entanglement, leading to degraded background consistency, weakened world memory, or conflicting motion-camera control.

To stabilize adaptation, we divide model parameters into \emph{slow memory} and \emph{fast memory}. The original self-attention layers and text-related cross-attention layers are treated as slow memory, since they encode the pretrained model's long-range spatiotemporal prior and should remain relatively stable. Newly introduced control modules, including $t$-guided projection layers, motion-control adapters, and camera-control branches, are treated as fast memory, as they need to rapidly absorb task-specific control information.

We implement this design with a differential learning-rate strategy:
\begin{equation}
    \theta_s \leftarrow \theta_s - \eta_s \nabla_{\theta_s}\mathcal{L},
    \qquad
    \theta_f \leftarrow \theta_f - \eta_f \nabla_{\theta_f}\mathcal{L},
\end{equation}
where $\theta_s$ and $\theta_f$ denote slow and fast parameter groups, respectively, with $\eta_s < \eta_f$.

This strategy preserves the pretrained autoregressive world prior while allowing control modules to adapt efficiently. By separating stable world-memory parameters from rapidly updated control parameters, Fast-Slow training improves optimization stability and reduces interference between heterogeneous control signals.

\subsection{Dataset Construction}
\label{sec:data_strategy}

\begin{figure}[t]
  \centering
  \includegraphics[width=1\linewidth]{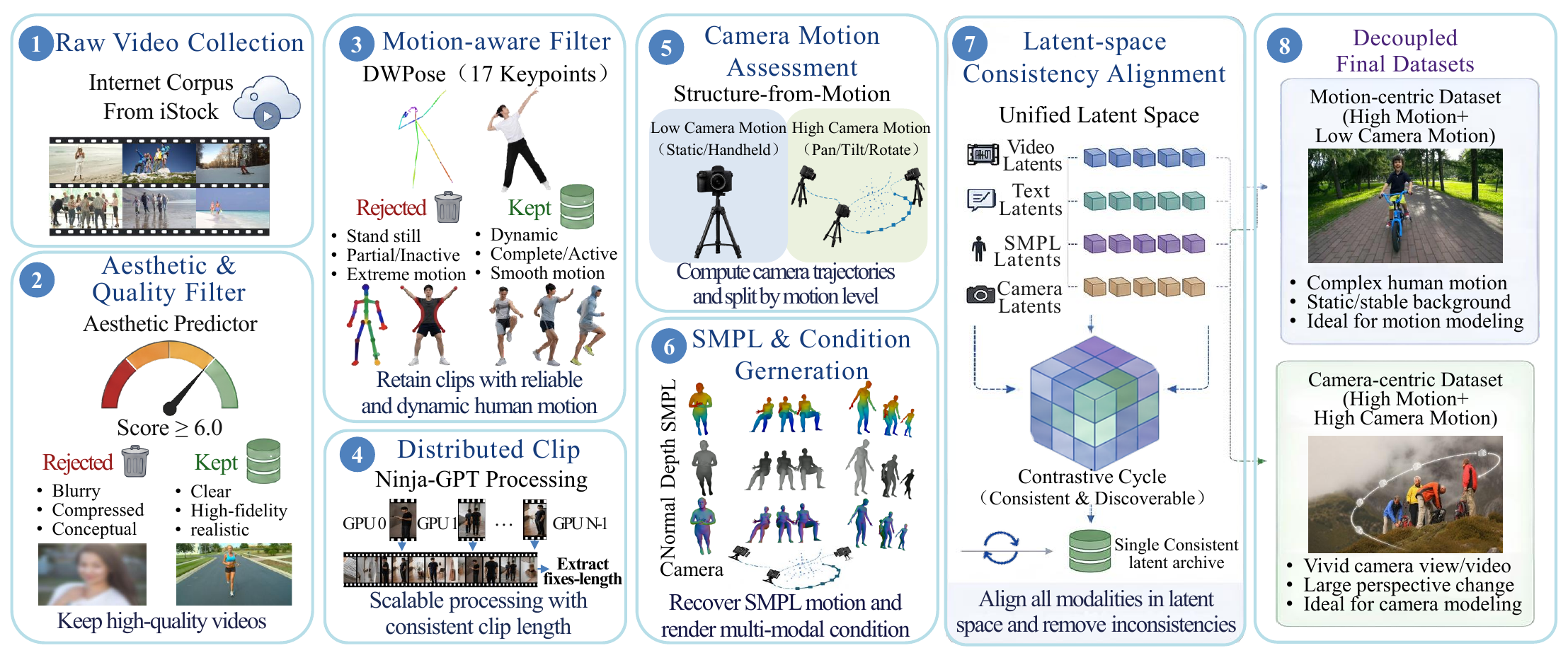}
\caption{
Dataset construction pipeline for two-stage controllable world-model training.
\textbf{1--4. Video Collection and Filtering.} We collect 20M iStock videos, remove low-quality samples with the LAION Aesthetic Predictor, retain clips with complete and dynamic human motion using DWPose, and extract fixed-length clips through distributed processing.
\textbf{5--7. Condition Generation and Alignment.} We estimate camera trajectories with Structure-from-Motion, recover SMPL motion, generate multi-modal conditions, and align video, text, SMPL, and camera conditions in a unified latent space.
\textbf{8. Decoupled Final Datasets.} The final data are split into a motion-centric subset for Stage-I human-motion control and a camera-centric subset for Stage-II world-exploration camera control.
}
  \label{fig:dataset}
\end{figure}

To support two-stage controllable world-model training, we construct a high-quality video-condition dataset from 20M publicly available stock videos crawled from iStock. Since the data are collected from publicly accessible stock media rather than private recordings, the dataset largely mitigates privacy concerns related to unauthorized filming or personal data collection. Since raw internet videos often contain low-quality content, incomplete humans, unreliable camera motion, or missing annotations, we design a filtering and alignment pipeline to retain clips with reliable human dynamics, valid camera trajectories, and synchronized multimodal conditions.

\noindent\textbf{Quality and Human-motion Filtering.}
We first remove corrupted or visually degraded samples using the LAION Aesthetic Predictor, discarding videos with $S_{aes}<6.0$. We then apply DWPose~\cite{dwpose} to extract 17 human keypoints and compute a motion score $S_{motion}$ based on pose completeness, motion range, and motion intensity. A frame is considered valid only when more than $70\%$ of joints are visible and core joints exceed a confidence threshold of $0.5$. We keep clips with complete and dynamic human motion while rejecting samples with excessive missing keypoints or unstable pose estimation, and extract fixed-length 81-frame clips from valid videos.

\noindent\textbf{Camera-motion Assessment.}
We estimate camera trajectories using Structure-from-Motion and compute a camera motion score $S_{camera}$ from translation magnitude and rotation variance. This score is used to distinguish low-camera-motion clips, which are suitable for motion-centric training, from high-camera-motion clips, which support camera-control learning. Such a decoupled data strategy prevents the model from learning complex human motion and large camera movement in a single entangled stage.

\noindent\textbf{Condition Generation and Alignment.}
For each retained clip, we use PromptHMR to estimate SMPL parameters and render geometry-aware condition maps, including canonical-normal maps, coordinate-space maps, and depth sequences. SfM-estimated camera trajectories are preserved as explicit camera-control conditions. We further encode video content, text captions, and SMPL conditions into a unified latent space, keeping only samples whose modalities are successfully generated and temporally synchronized.

\noindent\textbf{Two-stage Training Subsets.}
Based on $S_{motion}$ and $S_{camera}$, we build two disjoint subsets. The Stage-1 dataset $\mathcal{D}_{motion}$ contains clips with high $S_{motion}$ and low $S_{camera}$, enabling the model to learn SMPL-to-video control under relatively static backgrounds. The Stage-2 dataset $\mathcal{D}_{camera}$ contains clips with high $S_{motion}$ and high $S_{camera}$, enabling camera-control learning on top of the acquired human-motion prior. In total, our pipeline yields approximately 50k high-quality video-condition pairs, including 20k motion-centric samples and 30k camera-centric world-exploration samples.

\section{Experiments}
\label{sec:experiments}

\subsection{Experimental Setup}
\label{sec:exp_setup}

We evaluate our method on APRIL-AIGC/UltraVideo-Long~\cite{ultravideo} under motion-only, camera-only, and joint motion-camera control settings. Since different baselines support different types of control, we report their applicable settings together with the corresponding results. We randomly sample 100 long videos with durations from 15 seconds to 2 minutes, and uniformly extract DWPose and SMPL sequences as motion conditions. All methods are evaluated with their supported motion inputs and the same camera trajectories for fair comparison.

We assess generation performance from three aspects. First, following the standard VBench protocol, we evaluate visual and temporal quality using subject consistency, background consistency, aesthetic quality, imaging quality, temporal flickering, and motion smoothness. We report \textit{Consist} for subject/background consistency and \textit{Quality} for aesthetic/imaging quality. Second, we measure motion controllability by extracting DWPose keypoints from both the input condition and generated video, and computing the average displacement over 17 body keypoints. Third, we evaluate camera control with trajectory errors, including ATE, RPE, and RRE, which measure global trajectory deviation, local pose drift, and orientation error, respectively. Following common practice, camera trajectories are estimated by VGGT~\cite{vggt} and compared with the input camera conditions.

For ablation studies, we examine two core designs, Fast--Slow Memory Training and $t$-guided Projection. To reduce evaluation cost while keeping the same control setting, we conduct ablations on a 5-second test set under joint motion-camera control.

\begin{table*}[t]
\centering
\setlength{\tabcolsep}{5pt}
\renewcommand{\arraystretch}{1.15}
\resizebox{\linewidth}{!}{%
\begin{tabular}{l|ccc|ccccccccc}
\toprule
& \multicolumn{3}{c|}{Control}
& \shortstack{Overall\\Score}
& Consist
& Quality
& \shortstack{Temporal\\Flicker}
& \shortstack{Motion\\Smooth}
& \shortstack{Motion\\Error$\downarrow$}
& ATE$\downarrow$
& RPE$\downarrow$
& RRE$\downarrow$ \\
\cmidrule(lr){2-4}
Method
& Refined
& Motion
& Camera
&  &  &  &  &  &  &  &  &  \\
\midrule
Fun\-Camera~\cite{video_xfun}
&  &  & \checkmark
& /
& 86.59
& 59.77
& 96.70
& 98.75
& 0.213
& 0.601
& 0.101
& 0.970 \\

Fun\-Motion~\cite{video_xfun}
&  & \checkmark &
& /
& \cellcolor{red!25}91.76
&  62.52
& 97.40
& 98.71
& 0.184
& /
& /
& / \\

Wan\-Move~\cite{wan_move}
&  & \checkmark & \checkmark
& 3.401
& 89.67
& \cellcolor{red!25}64.23
& 96.67
& 98.22
& 0.194
& 0.669
& 0.077
& \cellcolor{yellow!25}0.685 \\

Uni3c~\cite{uni3c}
& \checkmark & \checkmark & \checkmark
& \cellcolor{yellow!25}3.445
& 88.65
& 56.07
& \cellcolor{yellow!25}97.94
& \cellcolor{yellow!25}99.05
& \cellcolor{red!25}0.151
& \cellcolor{red!25}0.482
& \cellcolor{red!25}0.063
& 0.722 \\

Ours
& \checkmark & \checkmark & \checkmark
& \cellcolor{red!25}3.868
& \cellcolor{yellow!25}90.25
& \cellcolor{yellow!25}62.98
& \cellcolor{red!25}98.62
& \cellcolor{red!25}99.35
& \cellcolor{yellow!25}0.161
& \cellcolor{yellow!25}0.532
& \cellcolor{yellow!25}0.070
& \cellcolor{red!25}0.349 \\
\bottomrule
\end{tabular}%
}
\caption{Quantitative comparison. ``Refined'' indicates whether the method provides fine-grained controllability beyond coarse trajectory-level guidance. Red and yellow cells denote the best and second-best scores, respectively. Some metrics are omitted for methods that do not support the corresponding control conditions.}
\label{tab:control_comparison}
\end{table*}

\subsection{Quantitative Results}
\label{sec:quantitative_results}

Table~\ref{tab:control_comparison} reports the quantitative comparison. The baselines support different control signals: \textit{Uni3C}~\cite{uni3c} uses both SMPL and camera parameters, \textit{FunCamera} uses camera parameters, \textit{FunMotion} uses DWPose-style skeletal conditions~\cite{dwpose}, and \textit{WanMove} supports only trajectory-level motion guidance. Our method also takes SMPL and camera parameters as input, targeting the challenging joint human-motion and camera-control setting. We therefore report results under each method's supported setting and compare both overall performance and control-specific metrics.

Among joint-control methods, our method achieves the best \textit{Overall Score}, showing a strong balance among visual quality, temporal stability, motion control, and camera control. This suggests that the proposed Fast--Slow Memory mechanism preserves the base model's generation capability while incorporating SMPL and camera conditions. Our method also performs well on \textit{Temporal Flickering}, \textit{Motion Smoothness}, and \textit{Motion Error}, indicating smoother temporal transitions and more faithful human-motion control. These gains are mainly brought by $t$-guided Dynamic Projection and Motion CFG, which improve SMPL condition injection across denoising stages. For camera control, our method achieves strong ATE, RPE, and RRE results, benefiting from the causal-aligned camera module and its integration with the MMPL-based autoregressive architecture~\cite{mmpl}. Although \textit{Uni3C}~\cite{uni3c} may slightly outperform on some fine-grained trajectory metrics, our method achieves better temporal consistency and overall performance, reflecting a more favorable trade-off between control accuracy and long-range coherence.

We also note that \textit{WanMove} achieves the best \textit{Quality} score and \textit{FunMotion} achieves the best \textit{Consist} score. However, these results are obtained under less demanding settings: \textit{WanMove} uses only coarse trajectory-level guidance, while \textit{FunMotion} does not involve camera control. In contrast, our method supports simultaneous SMPL and camera control, where maintaining visual quality, motion fidelity, and camera consistency is substantially harder. Overall, the results demonstrate that our method provides the most balanced performance under the joint-control setting.

\subsection{Qualitative Results}
\label{sec:qualitative_results}

\begin{figure}[t]
    \centering
    \includegraphics[width=1\linewidth]{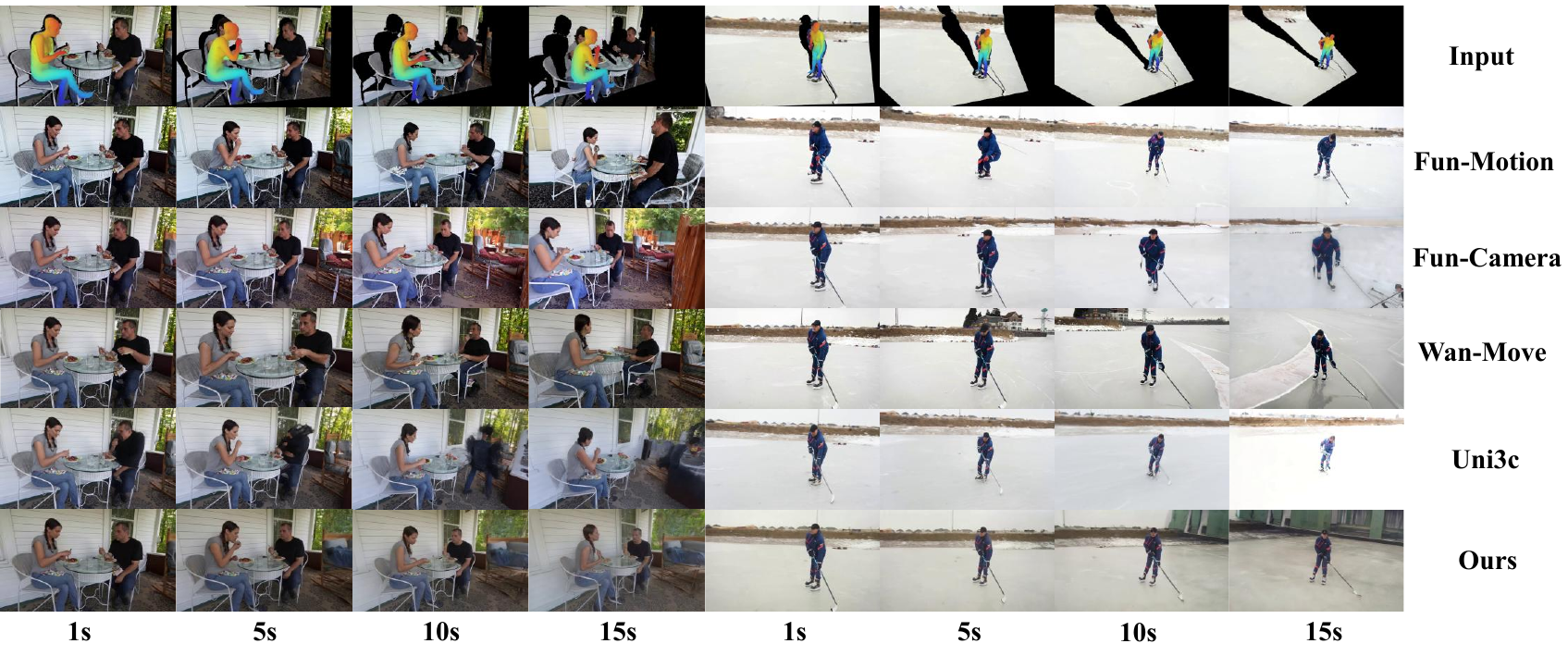}
    \caption{Qualitative comparison of long-video generation results on APRIL-AIGC/UltraVideo-Long. Our method produces more temporally coherent videos with better motion fidelity, stronger structural consistency, and more stable camera control over long horizons. For baseline methods that do not support dual control, we show results under their corresponding single-control setting.}
    \label{fig:qualitative_results}
\end{figure}

We further present qualitative comparisons to examine the visual behavior of different methods in long-video generation. Representative results are shown in Fig.~\ref{fig:qualitative_results}. Compared with competing approaches, our method produces videos with noticeably stronger temporal coherence, more stable subject appearance, and smoother motion transitions over long horizons.

A common failure mode of existing baselines is the gradual accumulation of temporal errors. In practice, this often manifests as identity drift, local structural distortions, background instability, and high-frequency flickering, especially in later segments of the generated video. These artifacts become even more pronounced in challenging scenarios involving large body motion, self-occlusion, or significant camera movement. By contrast, our method maintains globally consistent content and exhibits substantially improved robustness under these difficult conditions.

We further observe that single-control methods only address part of the world-modeling challenge. Motion-control methods can follow human poses in localized settings, but often lack world memory under large camera motion; camera-control methods support scene exploration, but do not provide precise control over active human dynamics. Thus, their advantages on individual metrics should be understood within their restricted settings. In contrast, our method targets the coupled problem emphasized in our introduction: preserving long-horizon world consistency while jointly controlling SMPL-based human motion and camera trajectories. By integrating both controls in a unified autoregressive world model, our method better supports interactive generation scenarios that require stable world memory, accurate camera exploration, and controllable human dynamics simultaneously.

Taken together, these qualitative results are consistent with the findings and further confirm the superiority of our method for stable and controllable long-video synthesis.

\subsection{Ablation Study}
\label{sec:ablation}

To better understand the contribution of each component in our framework, we conduct ablation studies on the key designs of our method, including the Fast-Slow Memory training strategy and the $t$-guided projection module.

\begin{figure}[t]
  \centering
  \includegraphics[width=1\linewidth]{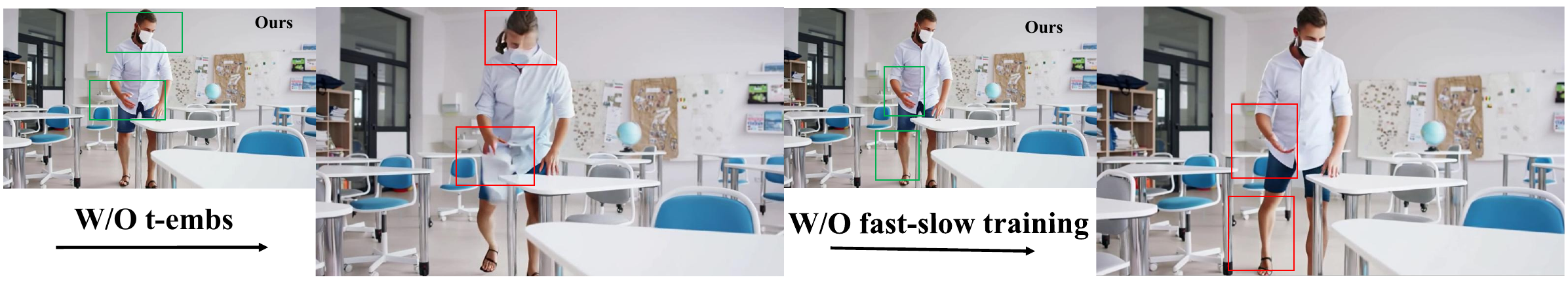}
   \caption{Ablation study on Fast-Slow Memory and $t$-guided projection. Without $t$-guided projection, motion control becomes overly strong, causing distorted poses, body tearing, and blurry artifacts. Without Fast-Slow Memory, the model struggles to preserve the pretrained generative prior, leading to degraded appearance and unstable body structure. Our full model better balances motion controllability and generation quality.}
   \label{fig:compar3}
\end{figure}

\noindent\textbf{Effect of Fast-Slow Memory Training.} As shown in Table~\ref{tab:fast_slow_memory_ablation}, Fast-Slow Memory Training consistently improves all evaluation metrics over direct fine-tuning.
In particular, it increases the overall score from 86.60 to 87.03, with clear gains in subject consistency, background consistency, aesthetic quality, imaging quality, and temporal stability.
These improvements indicate that directly updating all parameters can interfere with the pretrained long-video generative prior, leading to less stable optimization and weaker temporal consistency.
By separating slowly updated pretrained parameters from rapidly adapted task-specific parameters, Fast-Slow Memory preserves the original long-video prior while still allowing the model to learn the new control task effectively.
As a result, the model achieves better visual consistency, more stable temporal evolution, and stronger overall generation quality.

\begin{table}[t]
\centering
\footnotesize
\setlength{\tabcolsep}{6pt}
\renewcommand{\arraystretch}{1.15}
\begin{tabular}{l|ccccccc}
\toprule
Method
& Sub.
& Bg.
& Aes.
& Img.
& Temp.
& Motion
& Overall \\
\midrule
w/o Fast-Slow Memory Training
& 95.50
& 94.79
& 63.33
& 68.04
& 98.56
& 99.36
& 86.60 \\

w/ Fast-Slow Memory Training
& \cellcolor{red!25}96.18
& \cellcolor{red!25}95.30
& \cellcolor{red!25}64.30
& \cellcolor{red!25}68.37
& \cellcolor{red!25}98.65
& \cellcolor{red!25}99.37
& \cellcolor{red!25}87.03 \\
\bottomrule
\end{tabular}
\caption{Ablation study on Fast-Slow Memory Training. The overall score is computed as the average of all reported metrics.}
\label{tab:fast_slow_memory_ablation}
\end{table}

\noindent\textbf{Effect of $t$-guided Projection.} As shown in Table~\ref{tab:ablation_temb}, the proposed $t$-guided projection consistently outperforms the variants without timestep-aware condition adaptation.
Compared with naive condition injection, our method achieves the highest overall score and improves nearly all evaluation metrics, including subject consistency, background consistency, aesthetic quality, temporal flickering, and motion smoothness.
These results verify that motion conditions should be aligned with the denoising stage rather than injected in a timestep-agnostic manner.
Explicitly modeling this dependency allows the model to maintain stronger controllability and better generation stability throughout the denoising process.

\begin{table}[t]
\centering
\footnotesize
\setlength{\tabcolsep}{6pt}
\renewcommand{\arraystretch}{1.15}
\begin{tabular}{l|ccccccc}
\toprule
Method
& Sub.
& Bg.
& Aes.
& Img.
& Temp.
& Motion
& Overall \\
\midrule
w/o $t$-proj (6.5k)
& 95.38 & 94.52 & 63.01 & 67.08 & 98.21 & 99.13 & 86.22 \\

w/o $t$-proj (7k)
& 95.51 & 94.65 & 63.26 & 67.31 & 98.29 & 99.21 & 86.37 \\

w/o $t$-proj (8.5k)
& 95.65 & 94.83 & 63.07 & 68.50 & 98.29 & 99.27 & 86.60 \\

w/ $t$-proj (6.5k)
& \cellcolor{red!25}95.86 & \cellcolor{red!25}94.90 & \cellcolor{red!25}63.31 & \cellcolor{yellow!25}68.36 & \cellcolor{red!25}98.36 & \cellcolor{red!25}99.29 & \cellcolor{red!25}86.68 \\
\bottomrule
\end{tabular}
\caption{Ablation study on $t$-guided projection. The numbers appended to each method indicate the training steps of the evaluated checkpoint.}
\label{tab:ablation_temb}
\end{table}

Overall, the ablation study demonstrates that each component contributes positively to the final performance, while their combination yields the strongest results.

\section{Conclusion}
\label{sec:conclusion}

In this paper, we present \textit{Directing the World}, a unified autoregressive framework for controllable long-video generation with joint human-motion and camera-trajectory control. Built upon MMPL, our method introduces Fast-Slow Memory training for stable adaptation, a timestep-guided projection mechanism for SMPL-based motion control~\cite{smpl}, and a causal-aligned camera-control module compatible with block-wise autoregressive generation.

Experiments show that our framework achieves strong visual quality, temporal stability, motion fidelity, and camera controllability within a single autoregressive architecture. Qualitative results further demonstrate coherent subject appearance and smooth camera transitions over long horizons, even under large human and camera motions.

Several limitations remain. The current dataset scale and diversity still limit generalization to complex open-domain scenarios. Moreover, our method is less effective at fine-grained human dynamics, such as hand motion and facial expressions, especially when the subject is small. Future work may improve fine-grained controllability through larger motion-centric datasets and higher-resolution human representations.

Overall, we hope this work provides a useful step toward exploratory multi-agent world modeling and the evolution of controllable world simulation.

\clearpage
\appendix

\begin{center}
    \vspace*{1em}
    {\Huge \bfseries Supplementary Material \par}
    \vspace{1em}
\end{center}

\section{$t$-guided Projection Analysis}
\label{sec:supp_tguided_analysis}

We further analyze why directly injecting clean motion conditions into noisy diffusion latents can cause timestep-dependent instability, and why timestep-aware projection is necessary. Let
\[
z_t = \alpha_t z_0 + \sigma_t \epsilon,
\qquad
\epsilon \sim \mathcal{N}(0,I),
\]
denote the noisy latent at diffusion timestep $t$. Since $\epsilon$ is zero-mean and independent of the clean latent $z_0$, the cross term vanishes, and the second-order moment becomes $\mathbb{E}\|z_t\|_2^2 = \alpha_t^2 \mathbb{E}\|z_0\|_2^2 + \sigma_t^2 \mathbb{E}\|\epsilon\|_2^2$. This indicates that the latent statistics evolve continuously with timestep $t$: early denoising stages are dominated by stochastic noise, while later stages gradually approach the clean video manifold.

Now consider naive motion-condition injection, where $z_t^{\mathrm{naive}} = z_t + \lambda c_{\mathrm{smpl}}$ and $c_{\mathrm{smpl}}$ denotes the SMPL latent condition. Its second-order moment can be written as $\mathbb{E}\|z_t^{\mathrm{naive}}\|_2^2 = \alpha_t^2 \mathbb{E}\|z_0\|_2^2 + \sigma_t^2 \mathbb{E}\|\epsilon\|_2^2 + \lambda^2 \mathbb{E}\|c_{\mathrm{smpl}}\|_2^2 + 2\lambda\alpha_t \mathbb{E}\langle z_0,c_{\mathrm{smpl}}\rangle$, where the terms involving $\epsilon$ vanish due to independence and zero mean. Importantly, while the noisy latent statistics vary with $(\alpha_t,\sigma_t)$, the injected condition magnitude remains timestep-invariant. We therefore define the relative condition strength as
\[
\rho_t
=
\frac{
\lambda^2 \mathbb{E}\|c_{\mathrm{smpl}}\|_2^2
}{
\alpha_t^2 \mathbb{E}\|z_0\|_2^2
+
\sigma_t^2 \mathbb{E}\|\epsilon\|_2^2
}.
\]
Although $\rho_t$ is a simplified scalar measure and does not fully characterize the reverse denoising dynamics, it captures the key mismatch between a timestep-invariant condition and timestep-varying latent statistics.

At high-noise timesteps, the stochastic component dominates the latent, and the clean motion condition can be overwhelmed by noise. At low-noise timesteps, the stochastic component becomes weak and the latent lies closer to the clean video manifold. In this regime, the same timestep-invariant condition may act as an excessive perturbation to the local denoising trajectory, distorting appearance details, inducing motion jitter, or destabilizing the background.

To alleviate this mismatch, we replace the static condition with a timestep-aware projected condition:
\[
\tilde{c}_t = \mathcal{P}(c_{\mathrm{smpl}},t),
\qquad
z_t^{\mathrm{input}}
=
z_t + \lambda \tilde{c}_t .
\]
Accordingly, the relative condition strength becomes $\tilde{\rho}_t = \frac{\lambda^2 \mathbb{E}\|\tilde{c}_t\|_2^2}{\alpha_t^2 \mathbb{E}\|z_0\|_2^2 + \sigma_t^2 \mathbb{E}\|\epsilon\|_2^2}$. The projection module $\mathcal{P}$ takes the denoising timestep as an additional input, allowing the model to learn timestep-dependent rescaling and redistribution of the control feature. This reduces the mismatch between static condition statistics and timestep-varying noisy latent statistics. As a result, motion guidance can remain effective during high-noise stages while avoiding destructive over-conditioning during low-noise stages, leading to more stable and precise long-horizon controllable video generation.

\section{More Results}

\begin{figure}[!h]
  \centering
  \includegraphics[width=\columnwidth]{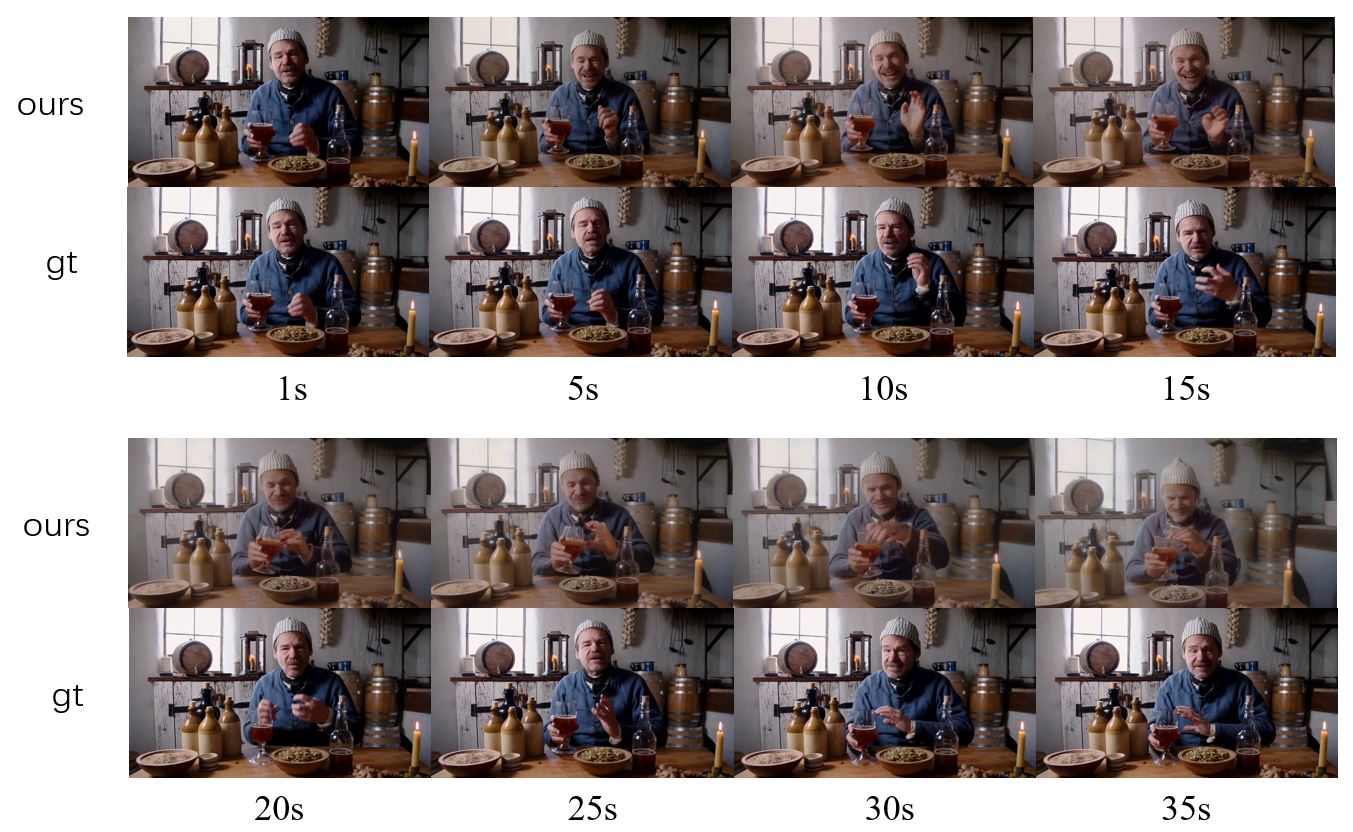}
    \caption{Comparison of long video generation results. Our method preserves subject identity and background consistency over time, but exhibits limitations in fine-grained motion control and temporal stability (e.g., brightness and color fluctuations).}
    \label{fig:longtime-generation}
\end{figure}

\paragraph{Qualitative Analysis.}
Figure~\ref{fig:longtime-generation} presents a comparison between our long video generation results and the ground truth over extended time horizons.
Our method demonstrates strong consistency in preserving both the primary human subject and the background content, indicating its ability to maintain global scene structure in long sequences.

\begin{figure}[!h]
  \centering
  \includegraphics[width=\columnwidth]{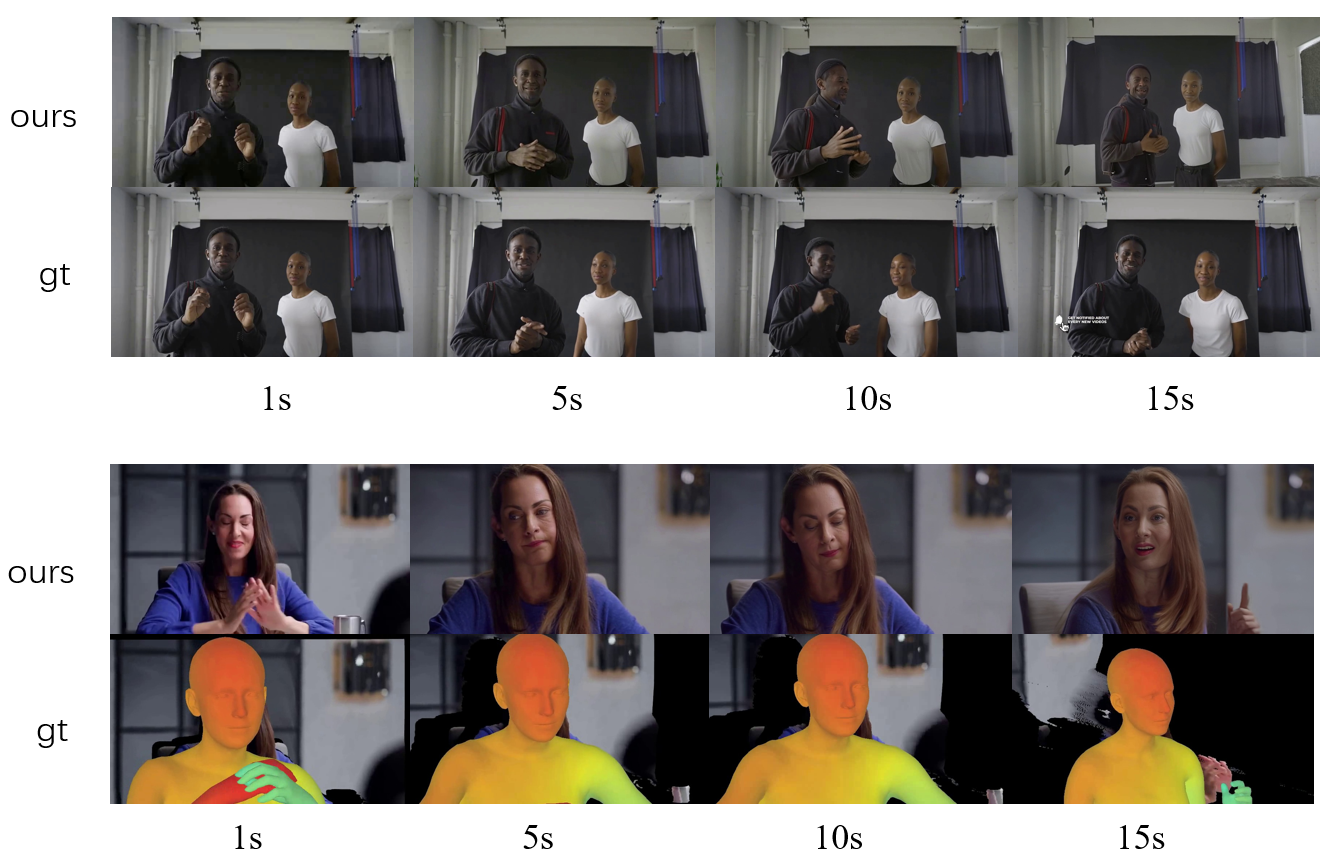}
    \caption{Single-person motion control and \textsc{SMPL} alignment results. Top: our method enables reliable control over a target subject even in multi-person scenes. Bottom: the generated results align well with \textsc{SMPL} guidance at the coarse motion level.}
    \label{fig:supp-2}
\end{figure}

Figure~\ref{fig:supp-2} further evaluates controllability.
As shown in the top row, our method enables reliable single-person motion control even in multi-person scenes, highlighting the effectiveness of our training strategy in isolating the target subject.
The bottom row shows that the generated results align well with the \textsc{SMPL} guidance at a coarse motion level, confirming that the model can follow structured motion inputs while maintaining reasonable visual fidelity.

\paragraph{Limitations.}
Despite these strengths, several limitations remain.

First, the model struggles with fine-grained motion control, particularly for subtle movements such as finger articulation.
This is likely due to two factors: (i) the limited granularity of \textsc{SMPL} estimation, and (ii) the loss of high-frequency motion details during patchification and 3D convolutional downsampling.
Consequently, the model tends to rely more on implicit motion priors rather than strictly adhering to precise control signals.

Second, error accumulation is still evident in long sequences, manifested as temporal inconsistencies such as brightness fluctuations and color shifts.
We hypothesize that this issue is related to the removal of text-prompt classifier-free guidance (CFG) and the introduction of motion-control CFG, which may weaken the model's ability to maintain stable global appearance.
A promising direction for future work is to incorporate text-prompt CFG distillation prior to integrating motion control, to better preserve appearance consistency.

Third, our camera control mechanism does not rely on explicit scene conditioning.
Unlike prior methods that render point cloud videos as input conditions---which often reduce the problem to an inpainting task---our approach directly controls camera motion via intrinsic and extrinsic parameters without injecting scene geometry or \textsc{SMPL} conditions into the input.
While this design improves scene consistency and avoids distortions introduced by imperfect point cloud rendering, it leads to slightly weaker accuracy in camera pose control compared to such baselines.
This suggests a trade-off between explicit scene conditioning and precise camera controllability.

\section{Training Details}
Our training pipeline consists of two consecutive stages: motion control training and camera control training.
In the first stage, we train the model with multi-person \textsc{SMPL} conditioning to learn robust human motion priors and capture complex motion dynamics, which requires approximately 650 H100 GPU hours.
In the second stage, we introduce a dedicated camera-control module to model viewpoint transformations and disentangle camera dynamics from human motion patterns, which requires around 1500 H100 GPU hours.
All experiments are conducted on 8 NVIDIA H100 GPUs with distributed training.
This two-stage training strategy allows the model to first establish reliable motion controllability and then further adapt to long-horizon camera control, leading to more stable and controllable video generation.

\section{Inference Pipeline}

During inference, our framework takes a single reference image, a reference motion video, and a camera trajectory as inputs. The image provides the target appearance and scene context, the motion video specifies the desired human dynamics, and the camera trajectory defines the intended viewpoint changes.

We first extract an SMPL video from the reference motion video as the human-motion control signal. Since this SMPL video is defined in the coordinate frame of the reference video, we align it to the target person in the input image. Specifically, we detect the target human region, extract its 17 COCO keypoints, and match them with the corresponding keypoints from the rendered SMPL body. Based on these keypoints, we estimate a lightweight global transformation, including rotation, translation, and scale, and apply it consistently to all SMPL frames. This produces an aligned SMPL control video that preserves the reference motion while matching the target subject in the input image. For multi-person control, we iteratively perform the same alignment for each target person and obtain multiple aligned SMPL control videos.

For camera control, we lift the input image into a coarse point cloud and render it along the prescribed camera trajectory, producing a trajectory-conditioned point-cloud video. This point-cloud sequence provides explicit geometric guidance for viewpoint changes. Finally, the aligned SMPL video, the input image, and the trajectory-conditioned point-cloud video are jointly used for autoregressive generation, enabling the model to follow both human-motion and camera-control signals while preserving appearance, identity, and scene coherence.

\section{Multiple-Person Motion Control}

A realistic world model should not only control a single human subject, but also support multiple active agents moving and interacting within the same persistent environment. To this end, our framework enables flexible multi-person motion control in image-conditioned video generation.

Given an input image with multiple people, our method supports \emph{targeted single-person control}, where a selected individual is driven by the reference SMPL motion while other subjects and the surrounding scene remain coherent. It also supports \emph{simultaneous multi-person control}, where different SMPL control videos are aligned to different target individuals and jointly used as motion conditions. In this way, multiple agents can follow their respective motion trajectories within the same generated world.

This capability is enabled by spatially aligned SMPL conditioning, where each control signal is associated with a specific person region in the input image. As a result, the model can reduce motion leakage across subjects and maintain identity consistency, temporal smoothness, and scene coherence. More importantly, multi-person control extends our framework from single-character animation toward controllable multi-agent world modeling, where coordinated human dynamics are generated as part of a shared evolving environment.

\begin{figure}[htbp]
    \centering
    \includegraphics[width=\linewidth]{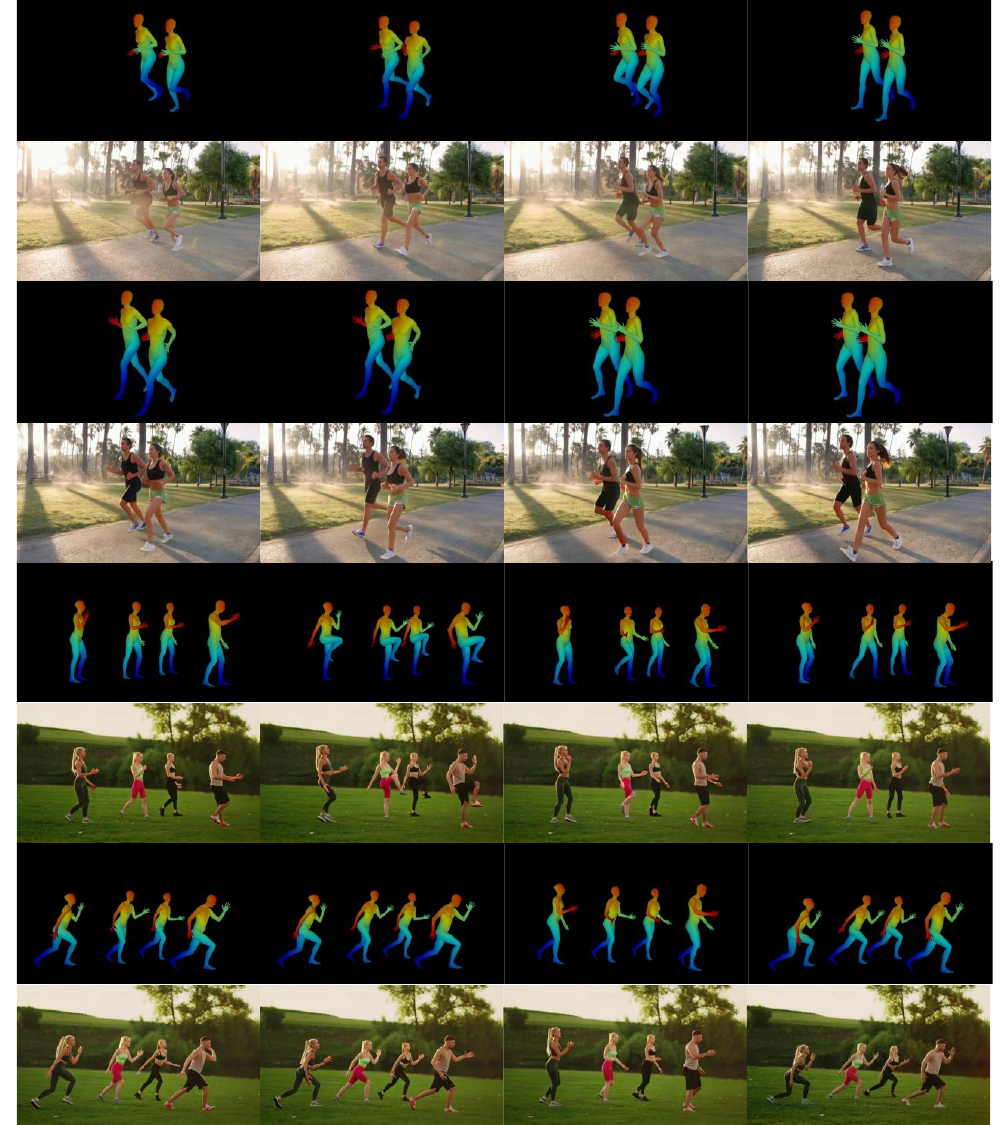}
    \caption{Multiple-person motion control}
    \label{fig:supp-3}
\end{figure}

\begin{figure}[htbp]
    \centering
    \includegraphics[width=\linewidth]{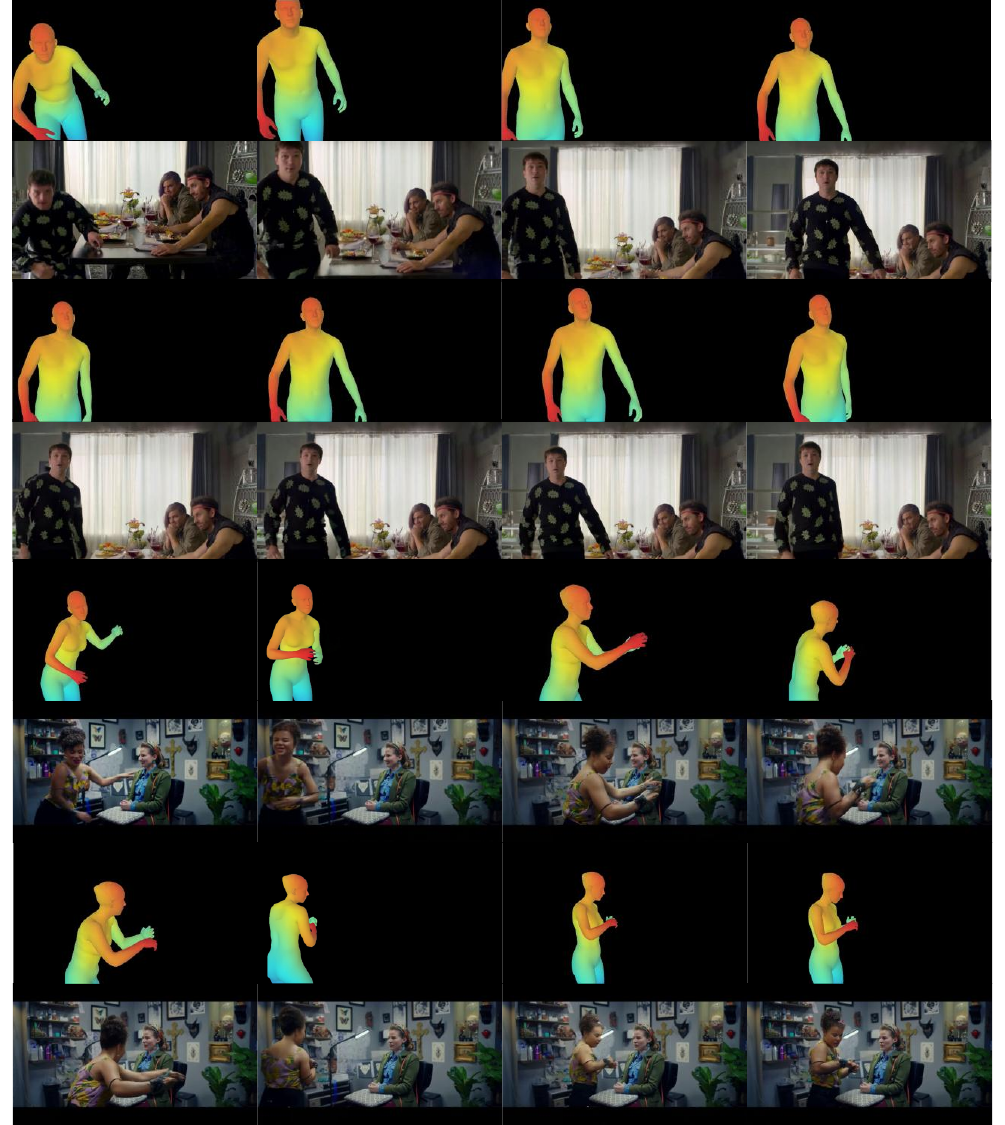}
    \caption{Single-person motion control with camera control}
    \label{fig:supp-4}
\end{figure}

\clearpage

\bibliographystyle{authordate1}
\bibliography{references}

\end{document}